\documentclass[letterpaper]{article} 
\usepackage[submission]{aaai24}  
\usepackage{xcolor}
\usepackage{times}  
\usepackage{helvet}  
\usepackage{courier}  
\usepackage[hyphens]{url}  
\usepackage{graphicx} 
\usepackage{natbib}  
\usepackage{caption} 
\usepackage{algorithm}
\usepackage{algpseudocode}
\usepackage{adjustbox}
\usepackage{multirow}
\usepackage{rotating}
\usepackage{amsmath}
\usepackage{newfloat}
\usepackage{listings}
\DeclareCaptionStyle{ruled}{labelfont=normalfont,labelsep=colon,strut=off} 
\floatstyle{ruled}
\newfloat{listing}{tb}{lst}{}
\floatname{listing}{Listing}
\graphicspath{{./Imgs/}}
\DeclareGraphicsExtensions{.pdf,.jpg,.png}

\title{Masked Diffusion with Task-awareness for Procedure Planning \\in Instructional Videos}
\author{
    Fen Fang\textsuperscript{\rm 1}, Yun Liu\textsuperscript{\rm 1}, Ali Koksal\textsuperscript{\rm 1}, Qianli Xu\textsuperscript{\rm 1}, Joo-Hwee Lim\textsuperscript{\rm 1,2}
}
\affiliations{
    \textsuperscript{\rm 1}Institute for Infocomm Research, Agency for Science, Technology, and Research (A*STAR), Singapore\\
    \textsuperscript{\rm 2}Nanyang Technological University, Singapore

    1 Fusionopolis Way, \#21-01, Connexis (South), Singapore, 138632\\
}

\begin{document}

\maketitle

\begin{abstract}
A key challenge with procedure planning in instructional videos lies in how to handle a large decision space consisting of a multitude of action types that belong to various tasks.
To understand real-world video content, an AI agent must proficiently discern these action types (\textit{e.g.}, \emph{pour milk}, \emph{pour water}, \emph{open lid}, \emph{close lid}, etc.) based on brief visual observation. Moreover, it must adeptly capture the intricate semantic relation of the action types and task goals, along with the variable action sequences.
Recently, notable progress has been made via the integration of diffusion models and visual representation learning to address the challenge. However, existing models employ rudimentary mechanisms to utilize task information to manage the decision space. To overcome this limitation, we introduce a simple yet effective enhancement - a masked diffusion model. The introduced mask acts akin to a task-oriented attention filter, enabling the diffusion/denoising process to concentrate on a subset of action types. Furthermore, to bolster the accuracy of task classification, we harness more potent visual representation learning techniques. In particular, we learn a joint visual-text embedding, where a text embedding is generated by prompting a pre-trained vision-language model to focus on human actions. We evaluate the method on three public datasets and achieve state-of-the-art performance on multiple metrics. Code is available at \url{https://github.com/ffzzy840304/Masked-PDPP}. 

\end{abstract}

\begin{figure}[!t]
\centering
\includegraphics[width=1.0\linewidth]{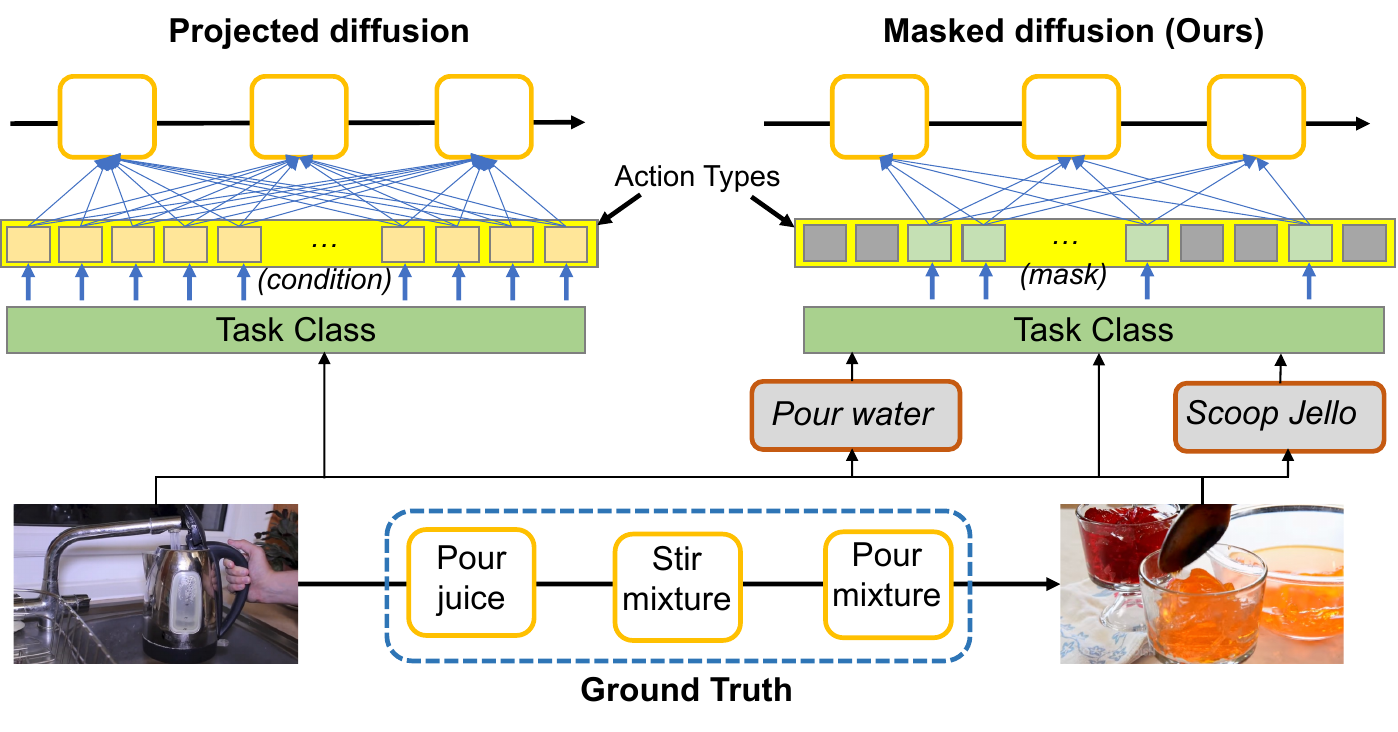}
\caption{
Searching and sorting action sequences from a large set of action types is challenging. 
Projected diffusion (top left) uses task class as a condition that does not restrict the decision space effectively. We propose masked diffusion (top right) to explicitly manage the decision space. Additionally, text embedding is used to enhance task classification and subsequent action sequence generation.}
\label{fig:illustration}
\vspace{-3mm}
\end{figure}

\section{Introduction}
Learning procedural knowledge from instructional videos - a natural ability of humans - presents a tough challenge to artificial intelligence (AI). It requires multiple aspects of cognitive and reasoning abilities such as scene understanding, event segmentation and discovery, action recognition and prediction, and causal reasoning \cite{Zhou2017TowardsAL, Zhou2023ProcedureAwarePF}. Building an AI agent with these capabilities is a pressing task for the AI community and has broad implications for real-world applications, \textit{e.g.}, to monitor human behaviors or to assist humans in collaborative tasks. In this paper, we focus on a sub-field of instructional video understanding, namely learning goal-directed actions from real-world videos, and subsequently generate feasible plans. In particular, we follow the work of \cite{Chang2019ProcedurePI} and cast the problem as procedure planning in instructional videos, which requires a model to generate action plans given the visual observations of a start state and a goal state (an example of \emph{making jello} is illustrated in Figure \ref{fig:illustration} - bottom). Moreover, we adopt the challenging problem setting of learning with weak supervision, \textit{i.e.}, to learn procedure knowledge without requiring intermediate visual observations \cite{Zhao2022P3IVPP}. Instead, only action labels 
are provided, which alleviates the costly annotation of the start and end times of each intermediate step. 

A key challenge with procedure planning in instructional videos lies in how to handle a large decision space consisting of a multitude of action types that belong to many tasks. For example, there are 778 action types from 180 task classes in the COIN dataset \cite{Tang2019-coin}. Since the datasets are collected from real-world videos at scale, the distribution of actions is largely unknown. In the current problem setting, the visual observations are essentially a pair of images (start and goal states) that are stochastically drawn from a video, and hence it is extremely difficult to recognize them from visual observations. Moreover, planning a sequence of actions from a large pool of actions is even more challenging considering the complicated semantic relationships between action types and task goals. This is exacerbated by the existence of multiple viable action plans to accomplish a specific task goal \cite{Bi2021ProcedurePI}.

Early works on procedure planning have employed a two-branch autoregressive approach while adopting different network architectures to model the probabilistic process. These include the dual dynamics networks (DDN) \cite{Chang2019ProcedurePI}, Bayesian Inference using exterior generative adversarial imitation learning (Ext-GAIL) \cite{Bi2021ProcedurePI}, and Transformers \cite{Sun2021PlaTeVP}. One limitation of these methods is related to the autoregressive process, which is slow and subject to error propagation. Moreover, they require the costly observation of intermediate states as supervisory signals. 
In contrast,  a single-branch non-autoregressive model is proposed that not only simultaneously predicts all intermediate steps, but more importantly alleviates the need for intermediate visual observations \cite{Zhao2022P3IVPP}. However, this method involves a complicated training process on multiple loss functions to manage a large design space. Recently, a diffusion-based probabilistic model is proposed to generate procedure plans non-autoregressively \cite{Wang2023PDPPPD}. It adopts a two-stage process, namely task classification and action sequence generation. The former aims to capture contextual information and used it as a conditional constraint in the latter step. However, as illustrated in Figure~\ref{fig:illustration} and shown by our results, using the task class as the condition has a limited effect on reducing the design space, \textit{i.e.}, decisions are still made with respect to a large pool of action types.


In this study, we propose a masked diffusion model to use task knowledge as context constraints. Instead of using task labels as a soft condition as in \cite{Wang2023PDPPPD}, we propose to generate a task-oriented mask to directly restrict the search space of action prediction. As shown in Figure \ref{fig:illustration}, action plans are generated on a greatly reduced subset of action types, owning to the task-guided mask. It helps to reduce the dimensionality of the decision space and enforces stronger hierarchical reasoning. Considering the possible adverse effect of inaccurate task classification, we further enhance visual representation learning via action-aware visual captioning based on pre-trained vision-language models (VLMs). In particular, a text embedding is obtained by prompting a frozen VLM (\textit{e.g.}, LLaVA) \cite{Liu2023VisualIT} to focus on the human actions in the current visual scene. We use text-enhanced multimodal embedding to both improve task classification and enhance action planning on the masked diffusion model. 

\textbf{Contributions}: (1) We propose a novel masked diffusion model to harness task information and enforce hierarchical procedure planning. Multiple strategies of masking operation are designed and evaluated to show the effectiveness of masking. (2) We enhance visual representation learning with an action-aware text embedding generated from a VLM in a zero-shot manner. We achieve state-of-the-art performance on multiple datasets under different testing conditions. These show the effectiveness of masked diffusion in planning under uncertainty and the potential of text-enhanced representation learning in procedure planning.

\section{Related Work}


\subsubsection{Action sequence modeling}
To handle complexities related to a large decision space, early works in procedure planning resort to solutions in probabilistic reasoning for goal-directed planning, such as universal planning networks \cite{UPN-Aravind}, uncertainty-aware action anticipation \cite{UAAA-Abu}, and causal InfoGAN \cite{Kurutach2018LearningPR}. However, these models have limited capacity in handling complexities in the scenes of instructional videos. The DDN model \cite{Chang2019ProcedurePI} learns the latent space via the interplay of a transition model and conjugate model, but suffers from compounding error. An Ext-GAIL model is proposed to separately handle time-invariant context knowledge and the casual relationship among actions, where a stochastic model is used to explicitly capture the distribution of plan variants during training \cite{Bi2021ProcedurePI}. The PlaTe model adopts transformer-based visual grounding and planning \cite{Sun2021PlaTeVP}, but has limited capacity to handle uncertainty. The above approaches suffer from slow convergence and error propagation owing to the auto-regressive reasoning process. 

Recently, a memory-enhanced probabilistic procedure planning framework is proposed, which adopts weak and adversarial supervision \cite{Zhao2022P3IVPP}. The method handles uncertainty by combining a pre-trained global memory unit, an adversarial generative model, and a Viterbi post-processing method. However, it involves a complicated training scheme and tedious inference process owing to the computation of multiple loss functions and the brittleness of training GANs. It is also restricted by the limited capability of a small-sized global memory with a fixed structure. 
The closest work to ours is the projected diffusion procedure planning (PDPP) model \cite{Wang2023PDPPPD}, which leverages the power of diffusion models to tackle complexity. However, task information is used as a ``soft'' condition in the representation, resulting in weak guidance to action planning. Moreover, task classification is performed using a simple multilayer perceptron (MLP) on standard visual embedding, which may not fully capture the value of the task context. 
We anticipate that context/task knowledge is crucial in effective and efficient procedure planning as is shown by numerous empirical evidences in hierarchical procedure planning \cite{Ashutosh2023HierVLLH,Nair2019HierarchicalFS,Liu2022LearningPT,Pertsch2020LongHorizonVP}. 

\subsubsection{Visual representation learning}
Visual reasoning can be enhanced by stronger visual representation learning. In the current problem formulation, the AI agent needs to infer the task type and generate action sequences based solely on two ``peeks'' into the start and goal states. 
Recently, notable progress has been made to train and fine-tune large VLMs \cite{Zhao2022LearningVR,Xu2021VideoCLIPCP,Lin2022LearningTR,Liu2023VisualIT}, which is partially driven by the availability of large-scale instructional video datasets \cite{Zhukov2019,Tang2019-coin, Damen2020RescalingEV, Miech2019HowTo100MLA,Grauman2021Ego4DAT}. The latest models usually use knowledge from the language domain (\textit{e.g.}, wikiHow) as distant supervision signals \cite{Zhong2023LearningPV,Zhou2023ProcedureAwarePF,Lin2022LearningTR}. However, the computational cost of training/fine-tuning large VLMs is usually prohibitively high. Alternatively, efforts have also been made to use pre-trained large language models (LLMs) as a visual planner \cite{Patel2023PretrainedLM,Wang2022MultimediaGS}, leveraging the zero-shot reasoning ability of powerful foundation models \cite{Ge2023ChainOT,Kim2022PureTA, OpenAI2023GPT4TR,Touvron2023LLaMAOA}. However, there is still a notable performance gap due to the lack of domain knowledge. 
Another stream of research resorts to graph-based representation to capture visual semantics of procedures, ranging from conventional neural task graph \cite{ Huang2018NeuralTG} to sophisticated transformer-based models \cite{Rampek2022RecipeFA,Mao2023ActionDT,Zhou2023ProcedureAwarePF}. One drawback of these models is that they are usually complex with an additional medium of graph representation.


\begin{figure*}[!t]
\centering
\includegraphics[width=1.0\linewidth]{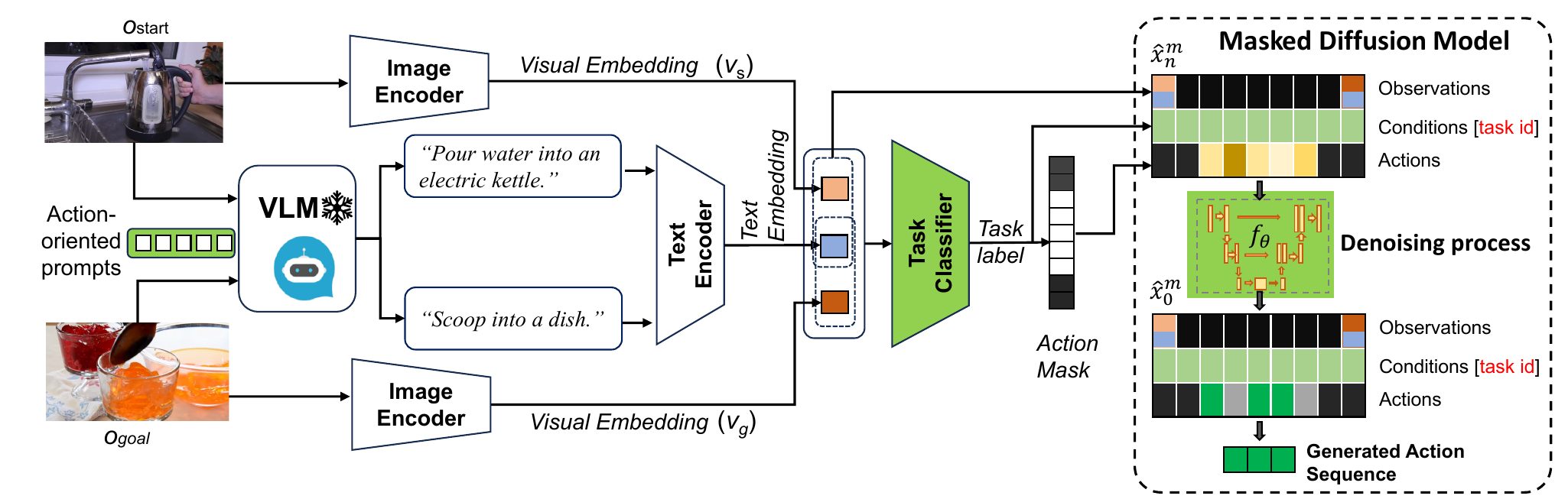}{}
\caption{Overview of our masked diffusion model with task-awareness. A frozen Visual-language model (VLM) generates text embedding of the start and goal states based on action-oriented prompts. An action mask is generated based on task class to restrict the action types.}
\label{fig:architecture}
\vspace{-3mm}
\end{figure*}

\section{Method}

\subsection{Problem Formulation}
Given the visual observations of a start state ($o_s$) and a goal state ($o_g$), the procedure planning task is to produce a plan in the form of a sequence of actions $a_{1:T}$ that, when executed, will enable the transition from $o_s$ to $o_g$ in $T$ steps, where $T$ is called the horizon of planning. Similar to \cite{Wang2023PDPPPD}, we decompose the task into two steps: (1) predicting the task category (\textit{e.g.}, \textit{make sandwich}, \textit{assemble bed}), and (2) generating action sequences conditioned on the predicted task category. The decision process can be formulated as 
\begin{equation}
p(a_{1:T}|o_s,o_g)= \int p(a_{1:T}|o_s,o_g, c)p(c|o_s,o_g)dc.
\label{eq:proc_plan}
\end{equation}
The system architecture is shown in Figure \ref{fig:architecture}.
As mentioned earlier, using task information as a condition does not exert sufficient modulation power on search space reduction. To address this issue, we propose a new strategy to make use of the task information, namely to generate a mask to restrict the decision space to a subset of ``promising'' actions. Notably, such a masking approach is 
different from masked diffusion transformers \cite{Zheng2023FastTO,Gao2023MaskedDT}. The latter aim to strengthen the model's ability to learn context information for image generation, whereas we use masks to restrict the decision space. 


\subsection{Action-aware Visual Representation Learning}
In \cite{Wang2023PDPPPD}, the task classifier is a simple MLP that takes the concatenated visual embedding of the start and goal state as input. In our model, task class plays an important role in the diffusion process by generating a mask to constrain the design space. Therefore, it is important to improve the accuracy of task classification. We propose two techniques to address this issue. First, we employ a Transformer model (\textit{e.g.}, ViT) (to replace the original MLP) that takes $o_s$ and $o_g$ (based on joint vision-text embedding) to predict the task class ($c$). Second, we enhance the visual representation by affixing an action-aware text embedding to the visual embedding. It is observed that prevalent image encoders are pre-trained on generic instructional videos, such as Howto100M \cite{Miech2019HowTo100MLA}, which does not possess sufficient ability to discriminate refined human actions. Fine-tuning such models is both costly and may jeopardize their generalisability. Meanwhile, numerous VLMs have been developed that show impressive descriptive power and flexibility. We adopt LLaVA \cite{Liu2023VisualIT} with frozen network weights\footnote{Other VLM models can be used to achieve similar outcome.} and prompt it to concentrate on the human actions in the visual input, \textit{e.g.}, ``\emph{Please briefly describe} [{\tt Image}] \emph{focusing on the human actions}''. A list of candidate prompts is included in the supplementary material. Despite the explicit request for brevity, the generated description may still be verbose and not suitable for subsequent reasoning. Therefore, we design a simple routine to extract the key words in the form of $verb + noun(s) + [optional] adverb(s)$. For example, in Figure \ref{fig:architecture}, the raw description of the start state can be ``\emph{In the image, a person is pouring water into an electric kettle from a faucet.}''. The routine can extract the key information, such as ``\emph{pour water into an electric kettle}''. Textual descriptions of $o_s$ and $o_g$ are fed into a pre-trained text encoder to generate the text embedding of the two states. Finally, the text embedding is concatenated with the visual embedding, resulting in the text-enhanced representation $(o^{VT}_s, o^{VT}_g)$. 

\subsection{Masked Diffusion Model}
In a standard diffusion model \cite{Ho2020DenoisingDP}, a forward diffusion process involves incremental addition of Gaussian noise $\epsilon \sim \mathcal{N}(0,\,\mathbf{I})$ to the input data ($x_0$, \textit{i.e.}, the true signal) until it degrades to a random Gaussian distribution $x_t$. The process is parameterized via a Stochastic Differential Equation (SDE):
%
\begin{equation} \label{eq:sde}
\begin{aligned}
q(x_n,x_0)= \sqrt{\overline{\alpha}_n}x_0+\epsilon\sqrt{1-\overline{\alpha}_n}, \\
q(x_n|x_{n-1})= \mathcal{N}(x_n;\sqrt{1-\beta_n}x_{n-1},\beta_n\mathbf{I}),
\end{aligned}
\end{equation}
where $\overline{\alpha}_n=\prod_{s=1}^n(1-\beta_s)$ denotes the noise magnitude, and $\beta_s\in(0,1)_{s=1}^t$ specifies the ratio of Gaussian noise added to the signal in each step. 

Similarly, the reverse denoising process gradually maps a Gaussian noise into the sample via a discrete SDE:
\begin{equation}
p_\theta(x_{n-1}|x_n)=\mathcal{N}(x_{n-1};\mu_\theta(x_n,n),\textstyle{\sum_\theta}(x_n,n)).
\label{eq:sde_3}
\end{equation}
The network is trained by optimizing the variational lower-bound of $p_\theta(x_0)$, based on which $\sum_\theta(x_n,n)$ can be obtained. Meanwhile, $\mu_\theta(x_n,n)$ is reparameterized as a noise prediction network $\epsilon_\theta(x_n,n)$, which is trained with a simple mean-squared error loss $L=||\epsilon-\epsilon_\theta(x_n,n)||^2$. After training, the model can recover the signal $x_0$ from random Gaussian noise.

We construct the input signal by concatenating three elements, namely (1) the text-enhanced visual observations of the start and goal states $(o^{VT}_s, o^{VT}_g)$, (2) the predicted task class ($c$), and (3) a sequence of candidate actions $(a_{1:T})$, i.e. $x=[(o^{VT}_s, o^{VT}_g),c,a_{1:T}]$. Different from \cite{Wang2023PDPPPD}, the candidate action sequence is affixed with a binary mask (\textit{e.g.}, `1' for active actions in the predicted task class, `0' for other actions) that is specified by the task class. In practice, the loss function is computed on $x$ with respect to the unmasked actions, $x^m$. The task-specific mask is derived from the mapping relationship between a task class and the action types, which can be obtained from ground truth during training. In essence, despite the fact that an individual action planning instance does not have the complete list of action types with respect to the task, one can simply include all action types for a specific task from many instances and remove the duplicates.

We adopt a similar condition project scheme on the task class and observations as in \cite{Wang2023PDPPPD}. Consistent with the premise that the initial and terminal actions are more important due to their primacy and recency effects, additional weights are assigned to these specific actions. The projection operation $Proj()$ in our model is defined as
\begin{equation}
\begin{matrix}    
\begin{bmatrix}
\hat{c}_1 & \hat{c}_2 & &\hat{c}_{T}\\
w\hat{a}_1^{m} & \hat{a}_2^{m} &... &w\hat{a}_T^{m} \\
\hat{o}_1& \hat{o}_2 & &\hat{o}_{T}
\end{bmatrix}&\rightarrow &\begin{bmatrix}
c & c & & c \\
w\hat{a}_1^{m} & \hat{a}_2^{m} &... &w\hat{a}_T^{m}\\
o_s^{VT} & 0& & o_g^{VT}
\end{bmatrix} \\
  x^m & &Proj(x^m) 
 \end{matrix},
 \label{eq:proj_func}
\end{equation}
where $\hat{c}_{i}$, $\hat{o}_{i}$ and $\hat{a}_{i}^{m}$ refer to the $i^{th}$ horizon task class, observation dimensions and predicted masked action logits in masked representation $x^m$, respectively. $c$,
$o_s^{VT}$, $o_g^{VT}$ represent the specified conditions.

The projection operation in Eq. \ref{eq:proj_func} indicates that the guidance is not changed during training. More importantly, after projecting task classification and observations to their original values, the loss on $x^m$ is exclusively attributed to $a^m$. Thus, the training loss can be computed as follows:
\begin{equation}
\mathcal{L}_{diff}^{m}=\sum_{n=1}^{N}(\epsilon_\theta(a_n^m,n)-a_0^m)^2.
\label{eq:training_loss}
\end{equation}
By employing a binary mask on the action dimensions, Gaussian noise is exclusively introduced to unmasked active actions. As a result, the search space for optimal actions is confined to the task-defined subset, rather than encompassing the entire action space of the dataset. This operation considerably reduces the learning load of the model during loss minimization, which in turn leads to a streamlined convergence process and enhanced accuracy in the denoising phase. This benefit becomes even more pronounced as the action space becomes larger.

\subsection{Training}
Our training program consists of two main stages: (1) training of a task class prediction model to extract conditional guidance from start to goal observation as well as action masks; (2) leveraging the masked diffusion model to effectively fit the target action sequence distribution. 

As mentioned earlier, a binary mask is applied to the action dimensions, directing the denoising model to focus on active actions. 
In the action sequence distribution fitting stage, we adopt the U-Net architecture \cite{Unet-Olaf-2015} to learn the noise prediction model $\epsilon_\theta(x_n,n)$ on the masked action distribution, as it resembles the stacked denoising autoencoders. By minimizing $\mathcal{L}_{diff}^m$, the model effectively mitigates the impact of randomly introduced noise on $x_n^m$. The detailed denoising model training process is shown in Algorithm \ref{alg:training}.

\begin{algorithm}[!t]
\caption{Training Process}\label{alg:training}
\begin{algorithmic}
\State \textbf{Input:} Initial input $x_0$, Gt task class $c$, the condition project function $Proj()$, total diffusion steps $N$, diffusion model $\epsilon_\theta$, $\{\overline{\alpha}_n\}_{n=1}^N$
\State 1: apply a binary mask to action dimension in $x_0$ given $c$ 
\State 2: $a_0^m=a_0[0,1,0...1,0]$($a$ in $c$ value is `1', otherwise `0')
\State 3: \textbf{repeat}
\State 4: \hskip1.0em $n\sim \{1,N\}$
\State 5: \hskip1.0em$\epsilon \sim \mathcal{N}(0,\,\mathbf{I})$
\State 6: \hskip1.0em $x_n^m$=$\sqrt{\overline{\alpha}_n}x_0^{m}+\epsilon\sqrt{1-\overline{\alpha}_n}$
\State 7: \hskip1.0em $\hat{x}_0^{m}=\epsilon_\theta(Proj(x_n^m),n)$
\State 8: \hskip1.0em Take gradient descent step on
\State 9: \hskip2.0em $\bigtriangledown_{  \theta}\parallel x_0^{m}-Proj(\hat{x}_0^{m}) \parallel ^2$
\State 10: \textbf{until} converged
\end{algorithmic}
\end{algorithm}
\vspace{-3mm}

\subsection{Inference}
During the inference stage, only the initial observation $o_s$ and the target observation $o_g$ are given. The task class is generated through the trained task classifier, eliminating the need for the ground truth task class as in the training phase. Subsequently, Gaussian noise is introduced to the conditions of the observations and masked action dimensions, resulting in the creation of $x_n^m$. The acquired denoise model is then employed to conduct denoising $N$ times for sampling an optimal action sequence. The detailed procedure in the inference stage is shown in Algorithm \ref{alg:inference}. 

\begin{algorithm}[!t]
\caption{Inference Process}\label{alg:inference}
\begin{algorithmic}
\State \textbf{Input:} Total diffusion steps $N$, task class prediction $c$, model $\epsilon_\theta$, $\{\overline{\alpha}_n\}_{n=1}^N$, $\{\beta_n\}_{n=1}^N$
\State 1: apply a binary mask to action dimension in $\hat{x}_N$ given $c$ 
\State 2: $\hat{a}_N^m=\hat{a}_N[0,1,0...1,0]$($a$ in $c$ value is `1', otherwise `0')
\State 3: \textbf{for} $n=N$,...,1 \textbf{do}
\State 4: \hskip1.0em $\hat{x}_0^{m}=\epsilon_\theta(Proj(x_n^m),n)$
\State 5: \hskip1.0em \textbf{if} $n>1$ \textbf{then}
\State 6: \hskip2.0em $\hat{\mu}_n=\frac{\sqrt{\overline{\alpha}_{n-1}}\beta_n}{1-\overline{\alpha}_n}\hat{x}_0^m+\frac{\sqrt{\alpha_n}(1-\overline{\alpha}_{n-1})}{1-\overline{\alpha}_n}\hat{x}_n^m$
\State 7: \hskip2.0em $\hat{\sum}_n=\frac{1-\overline{\alpha}_{n-1}}{1-\overline{\alpha}_n}\cdot \beta_n$
\State 8: \hskip2.0em $\hat{x}_{n-1}^m\sim \mathcal{N}(\hat{x}_{n-1}^m;\hat{\mu}_n,\hat{\sum}_n\mathbf{I})$
\State 9: \hskip1.0em \textbf{end if}
\State 10: \textbf{end for}
\State 11: return $\hat{x}_0^m$
\end{algorithmic}
\end{algorithm}

\subsection{Implementation Details}
 
The perceptual input to our model is a 1536-dimensional vector that represents the visual features extracted from HowTo100M \cite{Miech2019HowTo100MLA}. For the text representation input, we utilize LLaVA's \cite{Liu2023VisualIT} prompt-extracted text, which is subsequently encoded into a 578-dimensional vector using a DistilBERT \cite{Sanh2019DistilBERTAD} base model. All models are trained using a linear warm-up scheme. Throughout our experiments, the training batch size remains constant at 256. All the experiments are conducted using the ADAM optimizer \cite{DiederikADM} on a setup consisting of 4 NVIDIA RTX A5000 GPUs. Refer to supplement for more detailed information, such as learning rate and training epochs on different datasets.

\section{Experiments}
\subsection{Evaluation Protocol}

\subsubsection{Datasets} We conduct evaluations of our model on three instructional video datasets: CrossTask \cite{Zhukov2019}, NIV \cite{Alayrac16unsupervised}, and COIN \cite{Tang2019-coin}. 
The CrossTask dataset comprises 2,750 videos spanning 18 different tasks, with an average of 7.6 actions per video. The NIV dataset consists of 150 videos depicting 5 daily tasks, with an average of 9.5 actions per video. The COIN dataset contains 11,827 videos involving 180 different tasks, with an average of 3.6 actions per video. 
We adopt the standard approach by randomly splitting the data, using 70\% for training and 30\% for testing \cite{Zhao2022P3IVPP,Sun2021PlaTeVP,Wang2023PDPPPD}. We adhere to the data pre-processing methodology \cite{Wang2023PDPPPD} to generate action sequences and select \{start, goal\} observations.

\subsubsection{Metrics}
\label{sec:metrics}
In accordance with prior studies \cite{Zhao2022P3IVPP,Sun2021PlaTeVP,Wang2023PDPPPD}, we employ three metrics to assess the performance of our approach: 
(1) \emph{Success Rate (SR)}: A plan is considered correct only if all actions in the predicted sequence exactly match the corresponding actions in the ground truth. 
(2) \emph{Mean Accuracy (mAcc)}: It is the accuracy of actions at each individual time step. An action is considered correct if it precisely matches the action in the ground truth at the same time step. 
(3) \emph{Mean Intersection over Union (mIoU)}: It quantifies the overlap between predicted actions and the ground truth by computing the action IoU. Note that \emph{mIoU} does not consider the order of actions and solely indicates whether the model effectively captures the correct set of steps required to complete the procedure plan. Following \cite{Wang2023PDPPPD}, we calculate the \emph{mIoU} metric on each individual sequence, instead of computing it on every mini-batch, as done in prior studies \cite{Zhao2022P3IVPP,Sun2021PlaTeVP}. This is a more stringent condition and allows us to assess the accuracy of predicted actions for each specific sequence independently. 

In addition, we conduct a comprehensive evaluation of the stochastic nature of our model by employing various probabilistic metrics: (1) \emph{Kullback–Leibler divergence (KL-Div)} and \emph{Negative Log Likelihood (NLL)} between the probability distributions of the predicted plans and the corresponding ground truth; (2) \emph{Mode Recall (ModeRec)} to assess the coverage of ground truth modes in the results, and (3) \emph{Mode Precision (ModePrec)} to indicate the frequency with which our predicted plans align with the true modes of the data.

\subsubsection{Baselines}
We include recent procedure planning approaches based on instructional videos as baselines \cite{Chang2019ProcedurePI,Sun2021PlaTeVP,Bi2021ProcedurePI,Zhao2022P3IVPP,Wang2023PDPPPD}. 

\subsection{Task Classification Results} We intend to improve the task prediction accuracy by employing a combination of visual and text representations along with a transformer model. The results of task prediction performance are shown in Table \ref{tab:Task_classification}, where different configurations are examined. Our model achieves an improvement of approximately 3\% in task classification accuracy on the COIN dataset (with the largest task space). It  
achieves a slight improvement on CrossTask; and maintains perfect accuracy (100\%) on NIV as in other configurations. 

To verify the influence of task classification on the ultimate accuracy of action planning, we compare the outcomes achieved through the utilization of the MLP as detailed in \cite{Wang2023PDPPPD} with the results obtained by incorporating Transformer classifiers as inputs for both PDPP and our model. We observe a positive effect of Transformer, as shown by the results in supplementary section D.

\subsection{Comparison with Prior Approaches}
\subsubsection{Crosstask (short horizon)}
We show the main performance results on CrossTask in Table \ref{tab:crosstask}. Our model consistently outperforms other approaches in terms of both \emph{SR} and \emph{mAcc}. Across sequence lengths $T=3$ and $4$, our model exhibits a notable \emph{SR} increase of approximately 2\% (absolute change) compared to the previous state-of-the-art (SotA). In terms of \emph{mAcc}, our model showcases significant enhancements, achieving more than 11\% improvement at $T=3$ and around 2.3\% at $T=4$. Regarding \emph{mIoU}, as aforementioned, we follow PDPP \cite{Wang2023PDPPPD} to compute it by calculating the mean of every IoU for a single action sequence rather than a mini-batch adopted by \cite{Sun2021PlaTeVP,Zhao2022P3IVPP}. Hence, a direct comparison with \cite{Sun2021PlaTeVP,Zhao2022P3IVPP} is not relevant. Compared to PDPP, our model achieves about $1.5\%$ improvement in \emph{mIoU}.
\subsubsection{CrossTask (long horizon)} Following \cite{Zhao2022P3IVPP,Wang2022MultimediaGS}, we evaluate the performance on predicting plans for longer time horizons, $T={3,4,5, 6}$. The results are shown in Table \ref{tab:crosstask_long_path_planning}. Our model consistently achieves substantial enhancements across all planning horizons, surpassing the performance of previous models.

\subsubsection{NIV and COIN}
\begin{figure}[!t]
\centering
\includegraphics[width=0.85\linewidth]{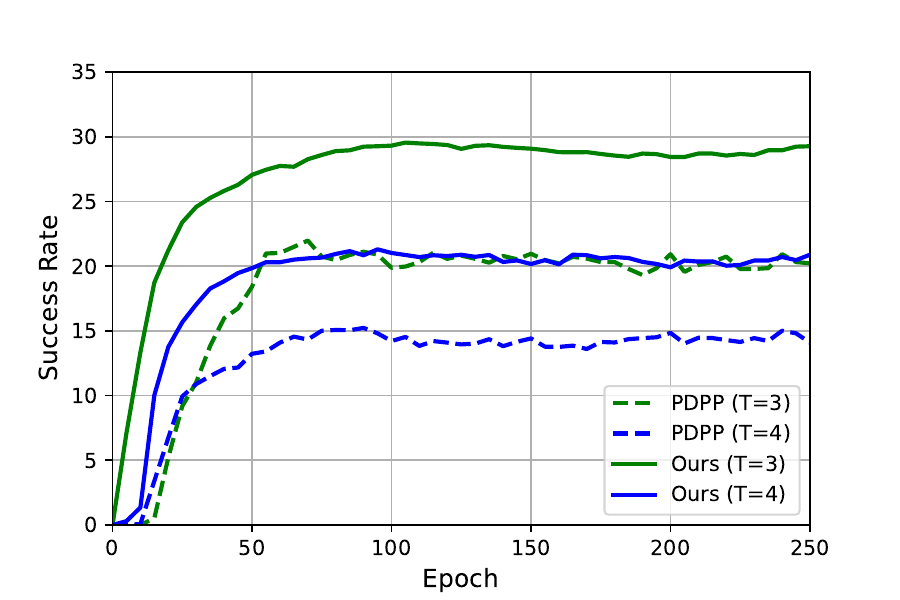}{}
\caption{Success rate during training on COIN dataset.}
\label{fig:plot}
\vspace{-3mm}
\end{figure}

Results on the NIV and COIN datasets are presented in Table \ref{tab:NIV_COIN}. It is shown that our method demonstrates superior performance on both datasets, surpassing other approaches in terms of \emph{SR} and \emph{mAcc} metrics. In particular, on the relatively smaller NIV dataset, our model achieves 1\% ($T=3$) and 2\% ($T=4$) increases respectively in \emph{SR}, along with improvements of 0.6\% ($T=3$) and 1.3\% ($T=4$) in \emph{mAcc}.
On the COIN dataset, which poses the highest level of difficulty, our method achieves a remarkable absolute improvement of 8.1\% ($T=3$) and 6.9\% ($T=4$) on \emph{SR}, and 4\% ($T=3$) and 2.7\% ($T=4$) on \emph{mAcc} metrics, respectively.
These represent a substantial margin over previous SotA, \textit{i.e.}, PDPP. Such a performance boost is also illustrated in Figure \ref{fig:plot}, which shows the training process on COIN dataset. Our approach features a large margin on SR and a faster learning speed, especially during the initial stage of training. 

\subsection{Evaluating Probabilistic Modeling}
To assess the effectiveness of our method on probabilistic modeling, we conduct a comparison between the plan distributions generated by our model and the ground truth distribution of viable plans, following the protocol proposed in \cite{Zhao2022P3IVPP,Wang2023PDPPPD}. The evaluation is done on CrossTask dataset which is most suitable for this purpose with its higher variations in feasible plans. Results on NIV and COIN datasets are included in the supplementary material.
We compare our model with three baselines: (1) a Deterministic baseline established by setting the initial distribution $\hat{x}_N = 0$, (2) a Noise baseline achieved by directly sampling from a random distribution using the provided observations and task class condition in a single step, and (3) the original PDPP approach \cite{Wang2023PDPPPD}.

The outcomes are presented in Table \ref{tab:crosstask_uncertainty}. Our model consistently produces the lowest \emph{NLL} and \emph{KL-Div} values across all horizons in comparison with the other three models. The results underscore the enhanced proficiency of our model in managing uncertainty. Furthermore, our model exhibits a remarkable capability to generate plans that are both diverse and logical, consistently outperforming the other models in terms of \emph{SR}, \emph{ModePrec}, and \emph{ModeRec} across all horizons.

\begin{table}[!t]
\centering
\caption{Task classification results. VM: visual representation + MLP classifier; VTM: visual-text representation + MLP classifier; VTT: visual-text representation + Transformer classifier.}
\begin{adjustbox}{width=0.45\textwidth}
\begin{tabular}{ccccc|cc|cc}
\hline
 & \multicolumn{4}{c}{CrossTask} & \multicolumn{2}{c}{COIN} & \multicolumn{2}{c}{NIV} \\\cline{2-9} 
\multicolumn{1}{l}{} & \textit{T=3} & \textit{T=4} & \textit{T=5} & \textit{T=6} & \textit{T=3} & \textit{T=4} & \textit{T=3} & \textit{T=4} \\ \hline
VM & 92.4 & 93.0 & 93.4 & 93.2 & 79.4 & 78.9 & 100 & 100 \\
VTM & 92.7 & 93.2 & 93.5 & 93.6 & 81.0 & 80.2 & 100 & 100 \\
VTT & \textbf{92.9} & \textbf{93.3} & \textbf{93.8} & \textbf{93.7} & \textbf{82.6} & \textbf{81.9} & \textbf{100} & \textbf{100} \\ \hline
\end{tabular}
\end{adjustbox}
\label{tab:Task_classification}
\end{table}

\begin{table*}[!ht]
\centering
\caption{Performance of benchmarks with planning horizons T$\in$\{3, 4\} on CrossTask. The `Supervision' column indicates the type of supervision during training. `V': intermediate visual states; `L': language features; `C': task class. Notably, to get \emph{mIoU}, we compute the average IoU for each individual action sequence, rather than across a mini-batch (in grey font).}

\begin{adjustbox}{width=0.85\textwidth}
\begin{tabular}{ccccc|ccc}
\hline
\multirow{2}{*}{Models} & \multirow{2}{*}{Supervision} & \multicolumn{3}{c}{\textit{T=3}} & \multicolumn{3}{c}{\textit{T=4}} \\ \cline{3-8} 
 &  & SR $\uparrow$ & mAcc $\uparrow$& mIoU $\uparrow$& SR $\uparrow$& mAcc $\uparrow$ & mIoU $\uparrow$ \\ \hline
DDN \cite{Chang2019ProcedurePI} & V & 12.18 & 31.29 & {\color{gray}47.48} & 5.97 & 27.10 & {\color{gray}48.46} \\
PlaTe \cite{Sun2021PlaTeVP} & L & 16.00 & 36.17 & {\color{gray}65.91} & 14.00 & 35.29 & {\color{gray}44.36} \\
Ext-GAIL \cite{Bi2021ProcedurePI} & V & 21.27 & 49.46 & {\color{gray}61.70} & 16.41 & 43.05 & {\color{gray}60.93} \\
P$^3$IV \cite{Zhao2022P3IVPP} & L & 23.34 & 49.46 & {\color{gray}73.89} & 13.40 & 44.16 & {\color{gray}70.01} \\
PDPP \cite{Wang2023PDPPPD} & C & 37.20 & 55.35 & 66.57 & 21.48 & 57.82 & 65.13 \\
Ours & C & \textbf{39.17} & \textbf{66.66} & 68.31 & \textbf{23.47} & \textbf{60.16} & 66.75 \\ \hline
\end{tabular}
\end{adjustbox}
\label{tab:crosstask}
\end{table*}

\begin{table}[!ht]
\centering
\caption{Results of success rate for longer planning horizons on CrossTask.}

\begin{adjustbox}{width=0.4\textwidth}
\begin{tabular}{ccccc}
\hline
\multirow{2}{*}{Models} & \textit{T=3} & \textit{T=4} & \textit{T=5} & \textit{T=6} \\ \cline{2-5} 
 & SR$\uparrow$ & SR$\uparrow$ & SR$\uparrow$ & SR$\uparrow$ \\ \hline
DDN \cite{Chang2019ProcedurePI} & 12.18 & 5.97 & 3.10 & 1.20 \\
PlaTe \cite{Sun2021PlaTeVP} & 18.50 & 14.00 & 10.00 & 7.50 \\
P$^3$IV \cite{Zhao2022P3IVPP} & 23.34 & 13.40 & 7.21 & 4.40 \\
PPDP \cite{Wang2023PDPPPD} & 37.20 & 21.48 & 13.58 & 8.47 \\
Ours & \textbf{39.17} & \textbf{23.47} & \textbf{15.25} & \textbf{10.10} \\ \hline
\end{tabular}
\end{adjustbox}
\label{tab:crosstask_long_path_planning}
\end{table}

\begin{table}[!ht]
\centering
\caption{Results for prediction horizons T$\in$\{3, 4\} on NIV and COIN datasets. `Sup.' means the type of supervision during training. }

\begin{adjustbox}{width=0.5\textwidth}
\begin{tabular}{cccccc|ccc}
\hline
\multirow{2}{*}{Hor.} & \multirow{2}{*}{Models} & \multirow{2}{*}{Sup.} & \multicolumn{3}{c}{NIV} & \multicolumn{3}{c}{COIN} \\ \cline{4-9} 
 &  &  & SR $\uparrow$ & mAcc $\uparrow$ & mIoU $\uparrow$ & SR $\uparrow$ & mAcc $\uparrow$ & mIoU $\uparrow$ \\ \hline
\multirow{5}{*}{\rotatebox{90}{\textit{T=3}}} 
 & DDN & V & 18.41 & 32.54 & {\color{gray}56.56} & 13.9 & 20.19 & {\color{gray}64.78} \\
 & Ext-GAIL & V & 22.11 & 42.20 & {\color{gray}65.93} & - & - & - \\
 & P$^3$IV & L & 24.68 & 49.01 & {\color{gray}74.29} & 15.4 & 21.67 & {\color{gray}76.31} \\
 & PDPP & C & 31.25 & 49.26 & 57.92 & 21.33 & 45.62 & 51.82 \\
 & Ours & C & \textbf{32.35} & \textbf{49.89} & 58.90 & \textbf{29.43} & \textbf{49.50} & 52.20 \\ \hline
\multirow{5}{*}{\rotatebox{90}{\textit{T=4}}} 
 & DDN & V & 15.97 & 27.09 & {\color{gray}53.84} & 11.13 & 17.71 & {\color{gray}68.06} \\
 & \multicolumn{1}{l}{Ext-GAIL} & V & 19.91 & 36.31 & {\color{gray}53.84} & - & - & - \\
 & P$^3$IV & L & 20.14 & 38.36 &{\color{gray} 67.29}& 11.32 & 18.85 & {\color{gray}70.53} \\
 & PDPP & C & 26.72 & 48.92 & 59.04 & 14.41 & 44.10 & 51.39 \\
 & Ours & C & \textbf{28.88} & \textbf{50.20} & 59.75 & \textbf{21.30} & \textbf{46.84} & 52.45 \\ \hline
\end{tabular}
\end{adjustbox}
\label{tab:NIV_COIN}
\end{table}

\begin{table}[!ht]
\centering
\caption{Uncertainty and diversity evaluation on CrossTask.}
\begin{adjustbox}{width=0.48\textwidth}
    \begin{tabular}{ccccccc}
\hline
Hori. & Model & NLL$\downarrow$ & KL-Div$\downarrow$ & SR$\uparrow$ & ModePrec$\uparrow$ & ModeRec$\uparrow$ \\ \hline
\multirow{4}{*}{\textit{T=3}} & Deterministic & 3.57 & 2.99 & 39.03 & 55.60 & 34.13 \\
 & Noise & 3.58 & 3.00 & 34.92 & 51.04 & 39.42 \\
 & PDPP & 3.61 & 3.03 & 37.20 & 53.14 & 36.49 \\
 & Ours & \textbf{3.10} & \textbf{2.52} & \textbf{39.17} & \textbf{56.03} & \textbf{44.52} \\ \hline
\multirow{4}{*}{\textit{T=4}} & Deterministic & 4.29 & 3.40 & 21.17 & 45.65 & 18.35 \\
 & Noise & 4.04 & 3.15 & 18.99 & 43.90 & 25.56 \\
 & PDPP & 3.85 & 2.96 & 21.28 & 44.55 & 31.10 \\
 & Ours & \textbf{3.41} & \textbf{2.52} & \textbf{23.47} & \textbf{47.33} & \textbf{34.60} \\ \hline
\multirow{4}{*}{\textit{T=5}} & Deterministic & 4.70 & 3.54 & 12.59 & 35.47 & 11.20 \\
 & Noise & 4.45 & 3.30 & 12.04 & 34.35 & 15.67 \\
 & PDPP & 3.77 & 2.62 & 13.58 & 36.30 & 29.45 \\
 & Ours & \textbf{3.61} & \textbf{2.46} & \textbf{15.25} & \textbf{36.56} & \textbf{30.12} \\ \hline
\multirow{4}{*}{\textit{T=6}} & Deterministic & 5.12 & 3.82 & 7.47 & 25.24 & 6.75 \\
 & Noise & 4.97 & 3.49 & 7.82 & 24.51 & 11.04 \\
 & PDPP & 4.06 & 2.76 & 8.47 & 25.61 & 22.68 \\
 & Ours & \textbf{3.67} & \textbf{2.37} & \textbf{10.10} & \textbf{25.90} & \textbf{28.69} \\ \hline
\end{tabular}
\end{adjustbox}
\label{tab:crosstask_uncertainty}
\vspace{-3mm}
\end{table}

\subsection{Ablation Studies}
\subsubsection{Effect of text-enhanced representation learning}
To validate the efficacy of text-enhanced representation learning within our model, we compare the performance of three setups: (1) the original PDPP model that utilizes only visual representation, (2) a truncated model that uses only text-based representation, and (3) a model that employs joint vision-text representations. The results are listed in Table \ref{tab:text-based representation.}. Apparently, the text-only modality is inferior to the visual-only modality and the vision-text multimodality in representation learning. Importantly, the additional action-aware text embedding does have a positive effect on planning efficacy as indicated by the higher performance of vision-text joint representation than visual-only. This outcome is consistent with the information in Table \ref{tab:Task_classification}, wherein higher accuracy of task classification is achieved when vision-text joint representation is used.

\begin{table}[!ht]
\centering
\caption{Ablation study on the role of text-enhanced representation learning.}
\begin{adjustbox}{width=0.48\textwidth}
\begin{tabular}{ccccc|cc|cc}
\hline
\multirow{2}{*}{Models} & \multicolumn{4}{c}{CrossTask} & \multicolumn{2}{c}{NIV} & \multicolumn{2}{c}{COIN} \\ \cline{2-9} 
 & \textit{T=3} & \textit{T=4} & \textit{T=5} & \textit{T=6} & \textit{T=3} & \textit{T=4} & \textit{T=3} & \textit{T=4} \\ \hline
PDPP(V) & 37.20 & 21.48 & 13.58 & 8.47 & 31.25 & 26.72 & 21.33 & 14.41 \\
PDPP(T) & 32.18 & 18.86 & 11.47 & 8.15 & 28.33 & 24.87 & 17.63 & 11.35 \\
PDPP(V+T) & 37.72 & 22.07 & 14.03 & 9.04 & 31.73 & 27.41 & 24.46 & 16.02 \\ \hline
\end{tabular}
\end{adjustbox}
\label{tab:text-based representation.}
\end{table}

\subsubsection{Effect of masked diffusion}
We conduct an ablation study to investigate the impact of different masking techniques on performance. In our method, we apply a binary mask to the action dimensions (called hard mask). Gaussian noise is exclusively generated within the unmasked regions, corresponding to the actions relevant to the active task. However, a possible adverse effect is that if the task classification is incorrect, there is a substantial likelihood that the action plans are wrong. Hence, we use the confidence score of task prediction to dictate the likelihood of a set of actions being retained, enabling the creation of a ``soft'' mask that is applied to the action dimensions. We also include a condition where no masking is applied to the action dimensions (w/o mask), resulting in a diffusion model identical to that of PDPP. 

The \emph{SR} results are outlined in Table \ref{tab:mask_ablation} - detailed data for \emph{SR}, \emph{mAcc}, and \emph{mIoU} can be found in supplementary materials. It is shown that hard masking results in the highest \emph{SR}. In fact, even without applying masking to action dimensions, the ``w/o mask" configuration outperforms PDPP, possibly due to improved task class prediction facilitated by text-enhanced representation. Interestingly, soft masking leads to the lowest \emph{SR}, performing worse than both PDPP and the non-masked approach. The possible reason is that with hard masking, action planning is confined within the boundaries of a task. This restriction significantly reduces the action space, allowing for a thorough exploration of the action sequencing within the unmasked subset of actions. With soft masking, the confidence scores of task classification could be ill-calibrated \cite{Guo2017OnCO}, which leads to wrong allocation to task-guided action types.

\begin{table}[!t]
\centering
\caption{Ablation study on the role of the masking type.}
\begin{adjustbox}{width=0.48\textwidth}
   \begin{tabular}{ccccc|cc|cc}
\hline
 & \multicolumn{4}{c}{CrossTask} & \multicolumn{2}{c}{NIV} & \multicolumn{2}{c}{COIN} \\\cline{2-9} 
 & \textit{T=3} & \textit{T=4} & \textit{T=5} & \textit{T=6} & \textit{T=3} & \textit{T=4} & \textit{T=3} & \textit{T=4} \\ \hline
PDPP & 37.20 & 21.48 & 13.58 & 8.47 & 31.25 & 26.72 & 21.33 & 14.41 \\ 
w/o mask & 37.72 & 22.07 & 14.03 & 9.04 & 31.73 & 27.41 & 24.46 & 16.02 \\
Soft mask & 34.37 & 18.44 & 12.04 & 7.73 & 30.44 & 26.07 & 18.76 & 12.57 \\
Hard mask & \textbf{39.17} & \textbf{23.47} & \textbf{15.25} & \textbf{10.10} & \textbf{32.35} & \textbf{28.88} & \textbf{27.85} & \textbf{20.24} \\ \hline
\end{tabular}
\end{adjustbox}
\label{tab:mask_ablation}
\end{table}

\section{Conclusion}
In this paper, we have introduced a masked diffusion model to deal with the large design space that challenges procedure planning in instructional videos. A simple yet effective masking mechanism is designed in a projected diffusion model to restrict the scope of planning to a subset of actions, as is guided by the task class information. We show that such a binary mask leads to significant improvements in procedure planning with respect to multiple metrics. It also engenders a positive effect on probabilistic modeling to reflect the inherent data distribution. Furthermore, we show the preferable effect of text-enhanced representation learning, which leverages the power of large VLMs and generates action-aware text description simply via prompting, without the need for computationally intensive training or fine-tuning. A direction of future work is to develop a more sophisticated masking scheme based on a well-calibrated task prediction model, so as to allow for a well-balanced compromise between the reduction in dimensions induced by masking and the retention of context relevant to the task.


\bigskip

\bibliography{aaai24}

\end{document}


\maketitle

\subsection{Supplementary Material Overview}
This supplementary material consists of the following contents. (1) In Sec. A, we provide the details of sub-models, including the task classification model at the first training stage and the diffusion model at the second stage. (2) In Sec. B, we describe the detailed baselines. (3) In Sec. C, we show the performance comparison using an alternative protocol on the CrossTask dataset \cite{Zhukov2019}. (4) In Sec. D, we present a comparison of the success rates in action planning on the COIN dataset \cite{Tang2019-coin}. We utilize various task classifiers to showcase the influence of task classification on the overall accuracy of the final action planning process. (5) In Sec. E, to demonstrate how the masking scheme facilitate decision making at varying sizes of the search space, we compare the performance gap between our model and PDPP on subsets of COIN that include different numbers of tasks (and hence action types). (6) In Sec. F, we provide more results on how our model handle uncertainty. (7) In Sec. G, we provide a list of prompts designed for the LLaVA model \cite{Liu2023VisualIT}. Additionally, we showcase textual descriptions of images generated by the visual-language model.

\subsection{A. Details of the sub-models in our method}
\label{sec:A}
\subsubsection{A.1 Transformer classifier}
\begin{figure}[!t]
\centering
\includegraphics[width=1.0\linewidth]{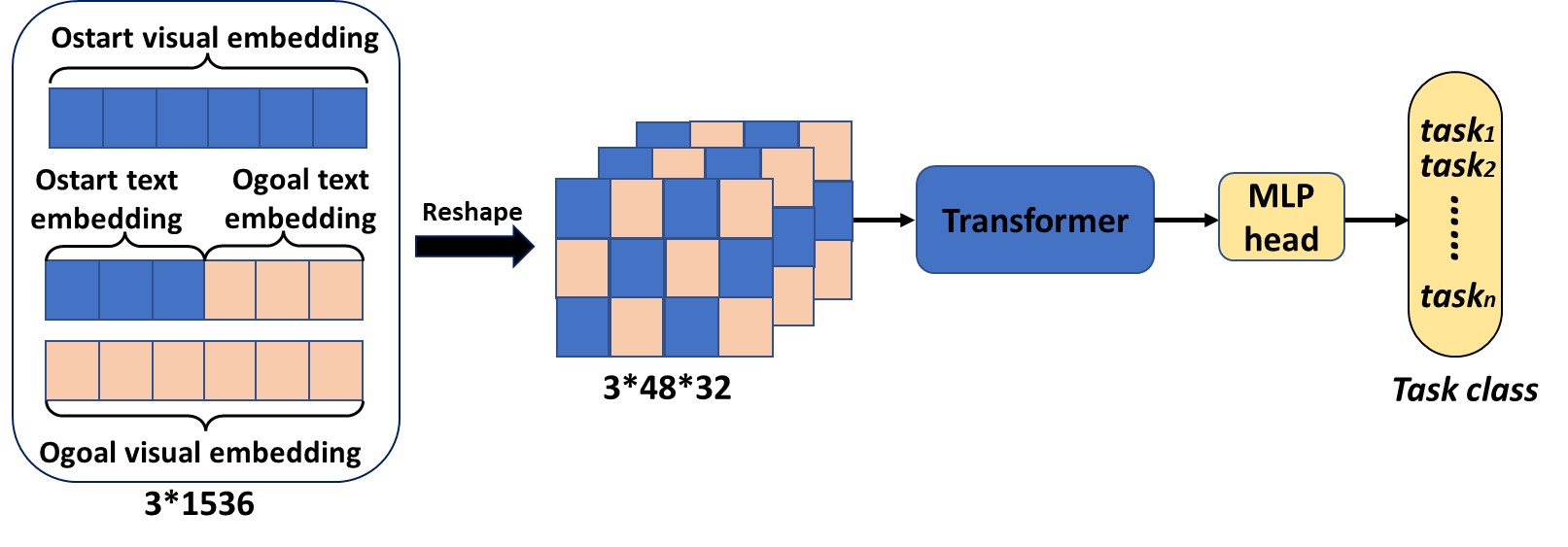}{}
\caption{
Task classifier architecture.}
\label{fig:task_classification}
\end{figure}

In the phase of task classification learning, we replace the MLP model \cite{Wang2023PDPPPD} with a Transformer, namely the ViT architecture \cite{dosovitskiy2020vit}. Our approach leverages visual representations denoted as a 1536-dimensional feature vector and text representations denoted as a 768-dimensional vector for both start and goal observations.
To configure the input for the ViT model, we amalgamate two text representations into a single 1536-dimensional feature. This combined feature is then triplicated and concatenated to create a 3-channel input, denoted as a $3\times1536$-dimensional feature array. These features are subsequently molded into a $3\times48\times32$ arrangement, optimally priming them for input into the ViT model.
Fig. \ref{fig:task_classification} gives a visual illustration of this procedure.

\subsubsection{A.2 Detail of diffusion model - U-Net}
We employ the popular U-Net \cite{Unet-Olaf-2015} architecture as the base diffusion model. Such an architecture resembles the stacked denoising autoencoders. Considering that the value of planning horizon is small ($T =$ \{3, 4, 5, 6\}), the U-Net is set as 3 layers. We configure the 1D-convolutional kernel with a size of 2, a stride of 1, and padding set to 0. This arrangement ensures that the length of the planning horizon dimension remains constant at 1 after each downsampling or upsampling operation.

The input to the diffusion model comprises the concatenation of task class, action labels, and observation features (including visual and text). As a result, the size of the feature dimension is denoted as $dim = L_c + L_a + L_o$, where $L_c$ refers to the number of task classes present in the dataset, $L_a$ represents the number of distinct actions within the dataset, and $L_o$ corresponds to the length of visual features. The feature dimensions in the three layers of the downsampling process evolve as follows: 256 $\rightarrow$ 512 $\rightarrow$ 1024. Conversely, in the upsampling process, the dimension transition is specified as 1024 $\rightarrow$ 512 $\rightarrow$ 256, ultimately restoring to the initial dimension, $dim$.

\subsection{A.3 Details of training process}
We employ a linear warm-up scheme during training, with slightly different settings to accommodate characteristics of individual datasets. 

For the CrossTask dataset \cite{Zhukov2019}, we configure a diffusion step value of 200. Our model undergoes a training process spanning 24,000 steps, with the learning rate progressively increasing linearly up to 5e-4 within the initial 4,000 steps. Subsequently, the learning rate undergoes a decay of 0.5 at the 10,000th, 16,000th, and 22,000th steps.

The NIV dataset \cite{Alayrac16unsupervised} is characterized by its smaller size. We set a diffusion step value of 50. The training duration spans 6,500 steps, during which the learning rate follows a linear increase up to 3e-4 for the first 4,500 steps, followed by a 0.5 decay at step 6,000.

For the COIN dataset \cite{Tang2019-coin}, we adopt a more extensive training regime due to its larger scale. We establish a diffusion step of 200 and a training duration of 160,000 steps. The learning rate commences with a linear increase to 1e-5 over the initial 4,000 steps, followed by 0.5 decay at the 14,000th and 24,000th steps. Subsequently, the learning rate is kept constant at 2.5e-6 for the remaining training iterations. 

In all experimental setups, the training batch size is set at 256. Our training process incorporates a weighted loss mechanism with a weight parameter (\textit{w}) set to 10. All experiments are conducted utilizing the ADAM optimization algorithm \cite{DiederikADM} on a setup involving 4 NVIDIA RTX A5000 GPUs.

\subsection{B. Baselines}
In this section, we present the baselines employed in our paper.
\begin{enumerate}
  \item [-]\textit{DDN} \cite{Chang2019ProcedurePI}. The DDN model comprises a dual-branch autoregressive structure, which is designed to acquire a conceptual representation of action steps and endeavors to prognosticate state-action transitions within the feature space.
  \item[-]\textit{PlaTe} \cite{Sun2021PlaTeVP}. The PlaTe model, similar to DDN, employs transformer modules in a dual-branch setup to facilitate its prediction process.
  \item [-] \textit{Ext-GAIL} \cite{Bi2021ProcedurePI}. This model addresses the task of procedure planning through reinforcement learning techniques. Similar to our approach, Ext-GAIL breaks down the procedure planning challenge into two distinct sub-problems. However, in Ext-GAIL, the primary objective of the first sub-problem is to supply extended horizon insights for the subsequent stage, whereas our primary intention is to establish conditions for the purpose of sampling.
 \item[-]\textit{P$^3$IV} \cite{Zhao2022P3IVPP}. P$^3$IV constitutes a transformer-based model operating within a single branch, enhanced by an adaptable memory bank and an additional generative adversarial framework. Similar to our approach, P$^3$IV also conducts simultaneous prediction of all action steps during the inference process.
 \item [-] \textit{PDPP} \cite{Wang2023PDPPPD}. PDPP formulates procedure planning as a distribution fitting problem. In this context, it characterizes the distribution of the complete intermediate action sequence using a diffusion model, effectively converting the planning problem into a sampling process from this distribution. It is worth noting that PDPP abstains from deploying resource-intensive intermediate supervision and instead leverages task labels sourced from instructional videos for guidance.
  \end{enumerate}
\subsection{C. Evaluation with another evaluation protocol}
Beyond the evaluation protocol used in the main paper (thereafter called ``Protocol 1''), previous works \cite{Sun2021PlaTeVP,Zhao2022P3IVPP,Wang2023PDPPPD} have reported results on an alternative evaluation approach referred to as ``Protocol 2". It features a different train/test split strategy and a different sampling method with respect to the planning horizon.  In particular, \textit{a}) ``Protocol 2" diverges in its data distribution, employing a partition of 2390 training samples and 360 testing samples. This distribution differs from ``Protocol 1'', which adopts a 70\%-train vs. 30\%-test split, affixed with a sliding window technique to derive the training data.
\textit{b}) In ``Protocol 2", one procedure plan with a prediction horizon $T$ is randomly chosen from each video for both training and testing purposes. This contrasts with ``Protocol 1'' that relies on a sliding window of size $T$ to encompass all procedure plans within each video.
\textit{c}) ``Protocol 2" modifies the prediction approach for a given planning horizon $T$, by restricting the prediction to encompass $T - 1$ actions, different from the prediction scope in ``Protocol 1''.

In the evaluation under ``Protocol 2," we conduct a comparative analysis of our model's performance against the results reported in prior studies on the CrossTask dataset. The comparison results are shown in Table \ref{tab:protocol2}. It is evident that our method consistently achieves the highest performance across all prediction horizons.

\begin{table}[!t]
\centering
\caption{Evaluation results of SR with protocol 2 on CrossTask. Prediction horizon is set to $T =$ \{3, 4, 5, 6\}.}

\begin{adjustbox}{width=0.35\textwidth}
\begin{tabular}{ccccc}
\hline
 & \textit{T=3} & \textit{T=4} & \textit{T=5} & \textit{T=6} \\ \hline
P$^{3}$IV & 24.40 & 15.80 & 11.80 & 8.30 \\
PDPP & 53.06 & 35.28 & 21.39 & 13.22 \\
Ours & \textbf{55.54} & \textbf{37.03} & \textbf{24.22} & \textbf{14.97} \\ \hline
\end{tabular}
\end{adjustbox}
\label{tab:protocol2}
\end{table}

\subsection{D. Influence of task classification}
As demonstrated in our main paper, task classification accuracy can be enhanced by substituting the MLP with a Transformer. We reported the performance of our model using the Transformer model as task classifier. To dissect the effect of task classifier (MLP vs. Transformer), we compare the final action planning success rates (\emph{SR}) using MLP and Transformer classifiers respectively within PDPP and our model in Table \ref{tab:mlp_transformer}. The evaluation is done on the COIN dataset only, considering that the effect of task classification could be weaker on CrossTask and NIV datasets, because task classification accuracy is already quite high on these datasets. It is evident that when the task classifier's performance improves, a corresponding enhancement can be observed in the final action planning \emph{SR} for both PDPP and our models.

\begin{table}[!t]
\centering
\caption{Comparision results (SR$\uparrow$) between using MLP and Transformer as the task classifier for PDPP and our model. The value in the bracket is the task classification accuracy.}
\begin{adjustbox}{width=0.45\textwidth}
\begin{tabular}{cccc}
\hline
Model & Classifier & $T=3$ & $T=4$ \\ \hline
\multirow{2}{*}{PDPP} & MLP & 21.33 (0.810) & 14.41 (0.802)\\
 & Transformer & 21.93 (0.826) & 15.03 (0.819) \\ \hline
\multirow{2}{*}{Ours} & MLP & 27.85 (0.810) & 20.24 (0.802)\\
 & Transformer & 29.43 (0.826) & 21.03 (0.819) \\ \hline
\end{tabular}
\end{adjustbox}
\label{tab:mlp_transformer}
\end{table}

\begin{figure}[!t]
\centering
\includegraphics[width=0.95\linewidth]{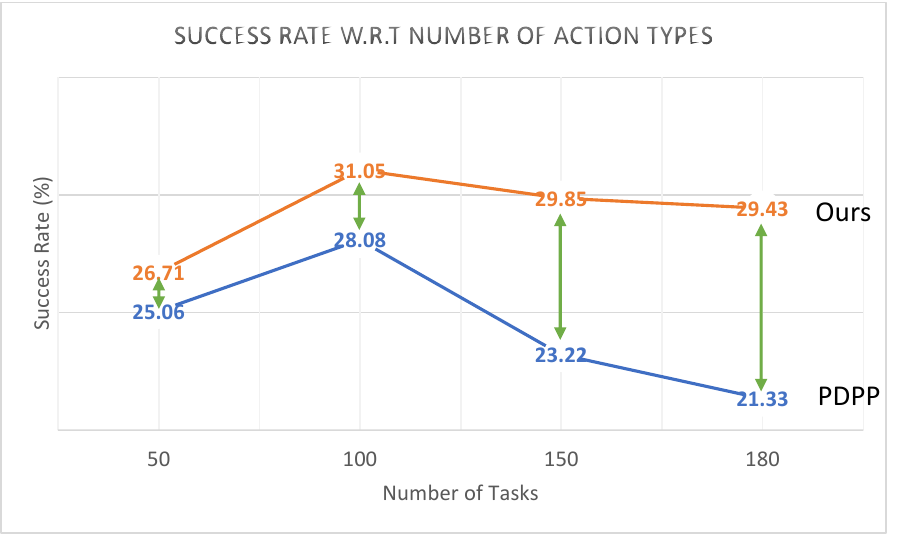}{}
\caption{Success rate comparison between PDPP and our method with different numbers of action types. Evaluated on the COIN dataset with the planning horizon $T=3$}
\label{fig:gap}
\end{figure}

\subsection{E. Influence of action searching space on procedure planning performance}
One motivation of our masked diffusion is to deal with the large search space embodied by many action types. We anticipate the benefit of masking is more profound when the search space is larger. To validate this hypothesis, we manipulate the size of search space (i.e., number of action types) and compare the performance (using $SR$ as the metric) of PDPP and our method. In our method, we use the MLP model for task classification as in PDPP to ensure that the performance differences are attributed to the usage of masked diffusion. In other words, we show if masking (our method) lead to higher performance gap compared to not using masking (PDPP) when handling larger search spaces. In implementation, we randomly extract 50, 100, and 150 tasks from the full COIN dataset (comprising 180 tasks) to create three distinct subsets. By increasing the number of tasks, the number of action types also increase. Subsequently, we subject these subsets to evaluation using both our model and PDPP. The $SR$ results of the two methods with a planning horizon of 3 are shown in Figure \ref{fig:gap}. It is evident that the performance advantage of our model over PDPP becomes more profound as the number of tasks/action types increase. When there are 50 different tasks, our model demonstrates a marginal performance improvement of just $1.65\%$ over PDPP. When the number of tasks increases to 100, the performance gain becomes $2.97\%$. Further extending the task number to 150, our model exhibits a substantial advantage with a margin of $6.63\%$. As shown in the main paper, our model outperforms PDPP by a significant margin of $8.10\%$ in terms of $SR$ on 180 task types on the full COIN dataset. 

Please note that this experiment aims to demonstrate the performance gap trend between our method and PDPP as the number of tasks changes. It is important to understand that a method's performance on various subsets might not consistently follow a trend due to potential biases introduced by randomly selecting tasks to form subsets. This helps explain the phenomenon that $SR$ of both methods shows improvement when applied to subsets with a task count ranging from 50 to 100. However, this trend reverses as the task count in subsets continues to increase to 100 and 150.


 
\subsection{F. Uncertainty modeling on NIV and COIN datasets}
In the main paper, we delve into an in-depth exploration of our model's probabilistic modeling prowess on the CrossTask dataset. Our investigation highlights the capacity of our masked diffusion model to generate plans that are not only accurate but also possess a commendable degree of diversity. In this document, we extend our analysis by offering comprehensive insights, presenting additional results of our model's ability to handle uncertainty.

We adhere to the protocol in \cite{Wang2023PDPPPD} to evaluate uncertainty and diversity. In the case of the Deterministic baseline, a single sampling process suffices to generate the plan, given that the outcomes are certain when observations and task class conditions are provided. For the Noise
baseline, the PDPP diffusion model, and our masked diffusion model, we sample 1,500 action sequences to derive our probabilistic outcomes for the computation of uncertainty metrics. Since the computational load is high, we employ the DDIM sampling method \cite{song2020denoising} to accelerate the sampling procedure within the PDPP diffusion model and our masked diffusion model.

We present the uncertainty modeling results on
NIV (Table \ref{tab:NIV_uncertainty}, Table \ref{tab:NIV_diversity_accuracy}) and COIN (Table \ref{tab:COIN_uncertainty}, Table \ref{tab:COIN_diversity_accuracy}), as an extension to that on the CrossTask dataset (presented in the main paper). The results on NIV show a striking similarity to those on CrossTask dataset. Specifically, both the PDPP and our approach exhibit a positive impact on the NIV dataset. In that sense, our method consistently demonstrates a superior capability in capturing uncertainty within procedure planning. The results also indicate that our method consistently excels in generating plans that exhibit a harmonious blend of diversity (measured by \emph{KL-Div and \emph{NLL}}) and rationality (measured by \emph{SR}, \emph{Prec} and \emph{Rec}) across both prediction horizons.

On the COIN dataset, it is reported (and replicated in our experiment) that the PDPP method had a detrimental effect on performance \cite{Wang2023PDPPPD}. This is indicated by the larger values of both Kullback-Leibler Divergence (\emph{KL-Div}) and Negative Log Likelihood (\emph{NLL}), which is inferior to the Deterministic method. Furthermore, the PDPP approach yields diminished values for Success Rate (\emph{SR}), mode Precision (\emph{Prec}), and mode Recall (\emph{Rec}) compared to the deterministic approach. The divergent performance of the PDPP method across datasets is assumed to stem from disparities in data scales and the inherent variability present in goal-conditioned plans within each dataset \cite{Wang2023PDPPPD}. In other words, given the substantial scale of the COIN dataset, the diffusion model encounters difficulties in effectively accommodating its intricacies. The introduction of noise to the model increases the difficulties of learning rather than alleviating them. In contrast, our masked diffusion model introduces noise solely within the confines of the task-defined subset. As a result, the search space for optimal actions is significantly narrowed when compared to the entirety of the dataset's action space.

\begin{figure*}[!t]
\centering
\includegraphics[width=0.95\linewidth]{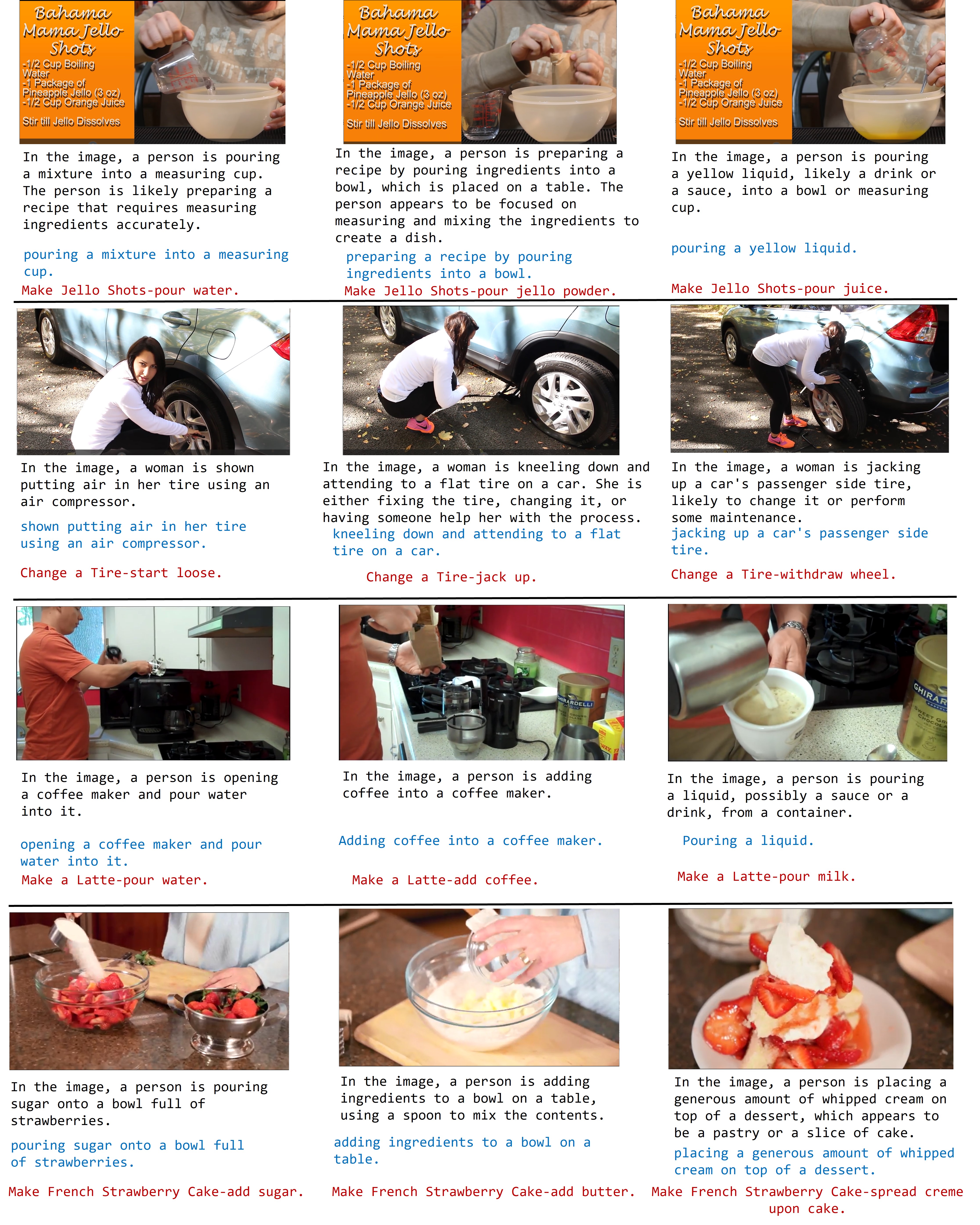}{}
\caption{Visualization of text-based action representations generated by LLaVA \cite{Liu2023VisualIT}. Black captions represent direct outputs generated by LLaVA using prompts, while blue captions denote extracted keywords, and red captions correspond to the ground truth task and action.}
\label{fig:vlm_samples}
\end{figure*}

As evidenced by the outcomes presented in Table \ref{tab:COIN_uncertainty}, our masked diffusion model exhibits remarkable proficiency in capturing the intricacies of uncertainty. Notably, our model boosts the lowest values for both \emph{KL-Div} and \emph{NLL} metrics across both horizons. Regarding diversity and accuracy, the results are shown in Table \ref{tab:COIN_diversity_accuracy}. In terms of Success Rate (\emph{SR}), our model demonstrates accuracy that is comparable to the Deterministic approach, with a marginal difference of $0.1\%$ lower at $T=3$, and a slightly superior performance at $T=4$. Furthermore, a substantial enhancement in \emph{SR} is evident in our model when compared to the PDPP model, with notable improvements of $6.6\%$ at $T=3$ and $5.8\%$ at $T=4$.  In terms of \emph{Prec}, our model achieves slightly lower values when compared to the Deterministic approach at both $T=3$ and $T=4$. However, our model still outperforms PDPP significantly, with a substantial margin of $4.74\%$ at $T=3$ and $5.51\%$ at $T=4$. For \emph{Rec}, our model surpasses the performance of the Deterministic approach by a substantial margin, with improvements of approximately $14\%$ at $T=3$ and $11.4\%$ at $T=4$. Furthermore, compared to PDPP, the advantages of our model become even more evident, with notable margins of around $18\%$ at $T=3$ and $15\%$ at $T=4$. A higher value of \emph{Rec} in our model indicates its ability to predict plans that encompass a greater number of ground truth modes.

\begin{table}[!t]
\centering
\caption{Evaluation results of the plan distributions metrics on
NIV.}

\begin{adjustbox}{width=0.45\textwidth}
\begin{tabular}{ccc|cc}
\hline
 & \multicolumn{2}{c}{\textit{T=3}} & \multicolumn{2}{c}{\textit{T=4}} \\ \cline{2-5} 
 & KL-Div$\downarrow$ & NLL$\downarrow$ & KL-Div$\downarrow$ & NLL$\downarrow$\\ \hline
Deterministic & 5.40 & 5.49 & 5.13 & 5.26 \\
Noise & 4.92 & 5.00 & 5.04 & 5.17 \\
PDPP & 4.85 & 4.93 & 4.62 & 4.75 \\
Ours & \textbf{4.49} & \textbf{4.56} & \textbf{4.54} & \textbf{4.51} \\ \hline
\end{tabular}
\end{adjustbox}
\label{tab:NIV_uncertainty}
\end{table}

\begin{table}[!t]
\centering
\caption{Evaluation results of diversity and accuracy metrics on
NIV.}

\begin{adjustbox}{width=0.48\textwidth}
\begin{tabular}{cccc|ccc}
\hline
 & \multicolumn{3}{c}{\textit{T=3}} & \multicolumn{3}{c}{\textit{T=4}} \\ \cline{2-7} 
 & SR$\uparrow$ & Prec$\uparrow$ & Rec$\uparrow$ & SR$\uparrow$ & Prec$\uparrow$ & Rec$\uparrow$ \\ \hline
Deterministic & 27.94 & 29.63 & 27.44 & 25.43 & 26.64 & 24.08 \\
Noise & 25.73 & 26.87 & 38.37 & 22.84 & 23.05 & 31.89 \\
PDPP & 31.25 & 31.78 & 33.09 & 26.72 & 29.10 & 33.08 \\
Ours & \textbf{32.35} & \textbf{31.93} & \textbf{40.58} & \textbf{28.88} & \textbf{29.63} & \textbf{38.09} \\ \hline
\end{tabular}
\end{adjustbox}
\label{tab:NIV_diversity_accuracy}
\end{table}

\begin{table}[!t]
\centering
\caption{Evaluation results of the plan distributions metrics on
COIN.}

\begin{adjustbox}{width=0.45\textwidth}
\begin{tabular}{ccc|cc}
\hline
 & \multicolumn{2}{c}{\textit{T=3}} & \multicolumn{2}{c}{\textit{T=4}} \\ \cline{2-5} 
 & KL-Div$\downarrow$ & NLL$\downarrow$ & KL-Div$\downarrow$ & NLL$\downarrow$\\ \hline
Deterministic & 4.52 & 5.46 & 4.43 & 5.84 \\
Noise & 4.55 & 5.50 & 4.52 & 5.92 \\
PDPP & 4.76 & 5.71 & 4.62 & 6.03 \\
Ours & \textbf{4.03} & \textbf{4.98} & \textbf{3.92} & \textbf{4.32} \\ \hline
\end{tabular}
\end{adjustbox}
\label{tab:COIN_uncertainty}
\end{table}

\begin{table}[!t]
\centering
\caption{Evaluation results of diversity and accuracy metrics on
COIN.}

\begin{adjustbox}{width=0.48\textwidth}
\begin{tabular}{cccc|ccc}
\hline
 & \multicolumn{3}{c}{\textit{T=3}} & \multicolumn{3}{c}{\textit{T=4}} \\ \cline{2-7} 
 & SR$\uparrow$ & Prec$\uparrow$ & Rec$\uparrow$ & SR$\uparrow$ & Prec$\uparrow$ & Rec$\uparrow$ \\ \hline
Deterministic & \textbf{27.96} & \textbf{34.35} & 27.40 & 19.98 & \textbf{30.65} & 19.63 \\
Noise & 18.49 & 25.67 & 29.82 & 12.58 & 22.55 & 19.32 \\
PDPP & 21.33 & 28.03 & 23.49 & 14.41 & 24.83 & 16.26 \\
Ours & 27.85 & 32.77 & \textbf{41.11} & \textbf{20.24} & 30.34 & \textbf{31.05} \\ \hline
\end{tabular}
\end{adjustbox}
\label{tab:COIN_diversity_accuracy}
\end{table}

\subsection{G. Prompts designed for the VLM model}
In our experiments, we tried to extract image captions with emphasis on human action using a VLM model (LLaVA) \cite{Liu2023VisualIT}. The following prompts are randomly chosen given an image.
{\tt
\small
\begin{enumerate}
\item[-] Please describe human action in this image.
\item[-] What is the person doing in the image?
\item[-] What is the human action in this image?
\item[-] What is the purpose of the human in this image?
\item[-] What action is the person in the image currently engaged in?
\item[-] Can you infer the person's action from the image?
\end{enumerate}
}%
In Figure \ref{fig:vlm_samples}, we present visualized examples of text-based action representations generated by LLaVA, showcasing image-related extracted keywords alongside the corresponding ground-truth task-action pairs. From examples in the figure, noticeable resemblances can be observed either in terms of \textit{verbs} or \textit{nouns} between the ground-truth action and our extracted key words. Such semantic similarity could be translated into enhanced representation learning, which ultimately help to boost accuracy of action planning.

\bigskip

\bibliography{aaai24}